%% file: root.tex
\documentclass[letterpaper, 10 pt, conference]{ieeeconf}  

\makeatletter
\let\NAT@parse\undefined
\makeatother

\IEEEoverridecommandlockouts

\usepackage{graphics}
\usepackage{epsfig}
\usepackage{mathptmx}
\usepackage{times}
\usepackage{amsmath}
\usepackage{amssymb}
\usepackage{xcolor}
\usepackage{booktabs}
\usepackage[hidelinks]{hyperref}
\usepackage[numbers,sort&compress]{natbib}
\usepackage{array}
\usepackage{caption}
\usepackage{graphicx}
\usepackage{tikz}
\usepackage{subcaption}
\usepackage{url}
\usepackage{hyperref}
\usepackage[T1]{fontenc}
\usepackage[normalem]{ulem}
\usepackage{stfloats}

\newcommand{\simulationenv}{simulation environment}
\newcommand{\targetrealenv}{target real environment}
\newcommand{\targetrealscene}{target real scene}

\title{\LARGE \bf 
DriveEnv-NeRF: Exploration of A NeRF-Based Autonomous Driving Environment for Real-World Performance Validation
}

\author{
Mu-Yi Shen$^{1*}$, Chia-Chi Hsu$^{1*}$, Hao-Yu Hou$^{1*}$, Yu-Chen Huang$^{1*}$, Wei-Fang Sun$^{2}$, \\Chia-Che Chang$^{3}$, Yu-Lun Liu$^{4}$, and Chun-Yi Lee$^{1}$%
\thanks{$^{*}$Equal contribution}
\thanks{$^{1}$Elsa Lab, National Tsing Hua University}
\thanks{$^{2}$NVIDIA AI Technology Center, NVIDIA Corporation}
\thanks{$^{3}$MediaTek Inc.}
\thanks{$^{4}$National Yang Ming Chiao Tung University}
}

\begin{document}

\maketitle
\thispagestyle{empty}
\pagestyle{empty}
\input{sections/0.abstract}
\input{sections/1.introduction}
\input{sections/2.related_works}
\input{sections/3.methodology}
\input{sections/4.experimental_results}

\input{sections/5.discussion_and_conclusion}
\input{sections/6.acknowledgement}

\bibliographystyle{unsrtnat}
\bibliography{refs}

\clearpage

\begin{figure*}[!t]
    \centering
    \includegraphics[width=0.16\textwidth, height=0.08\textheight]{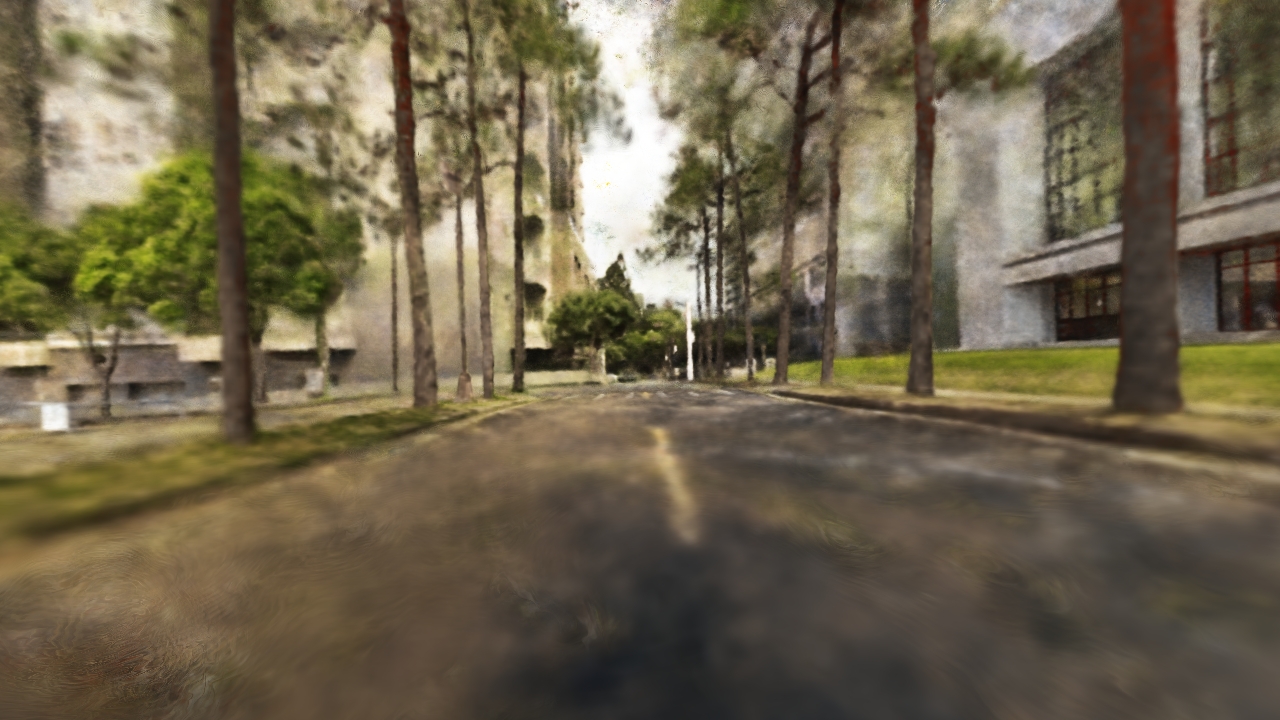}%
    \hfill
    \includegraphics[width=0.16\textwidth, height=0.08\textheight]{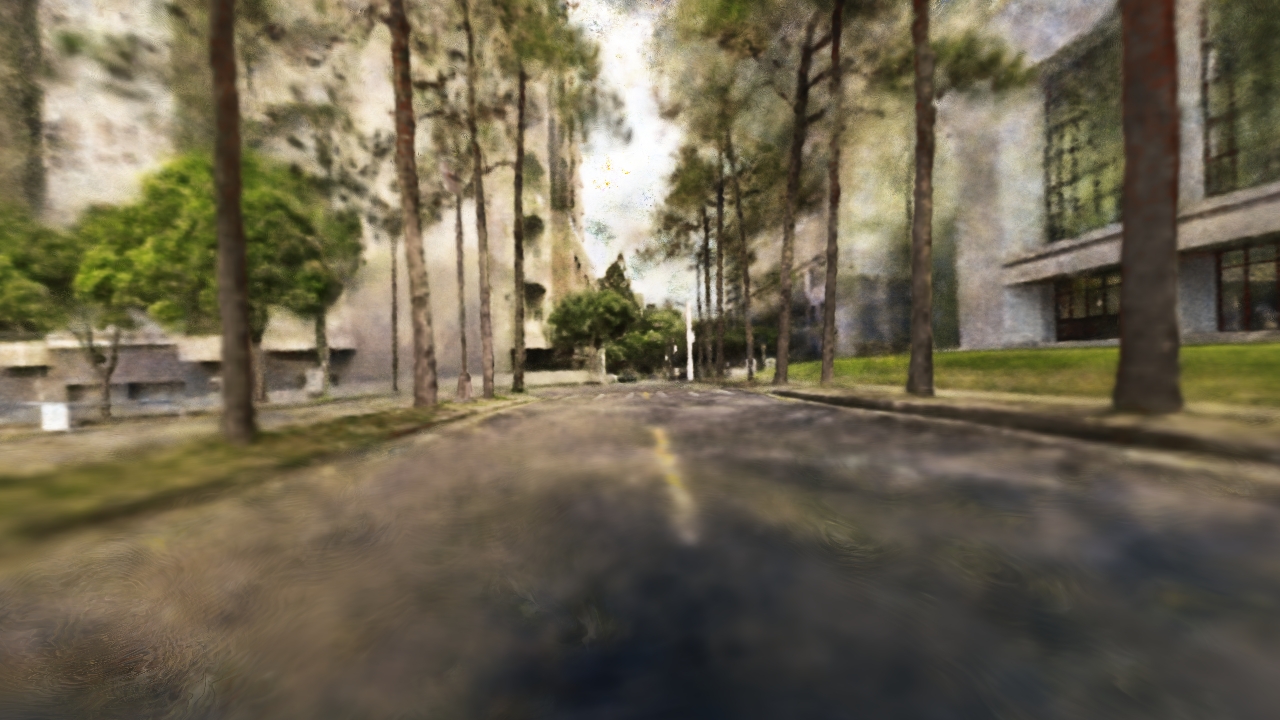}%
    \hfill
    \includegraphics[width=0.16\textwidth, height=0.08\textheight]{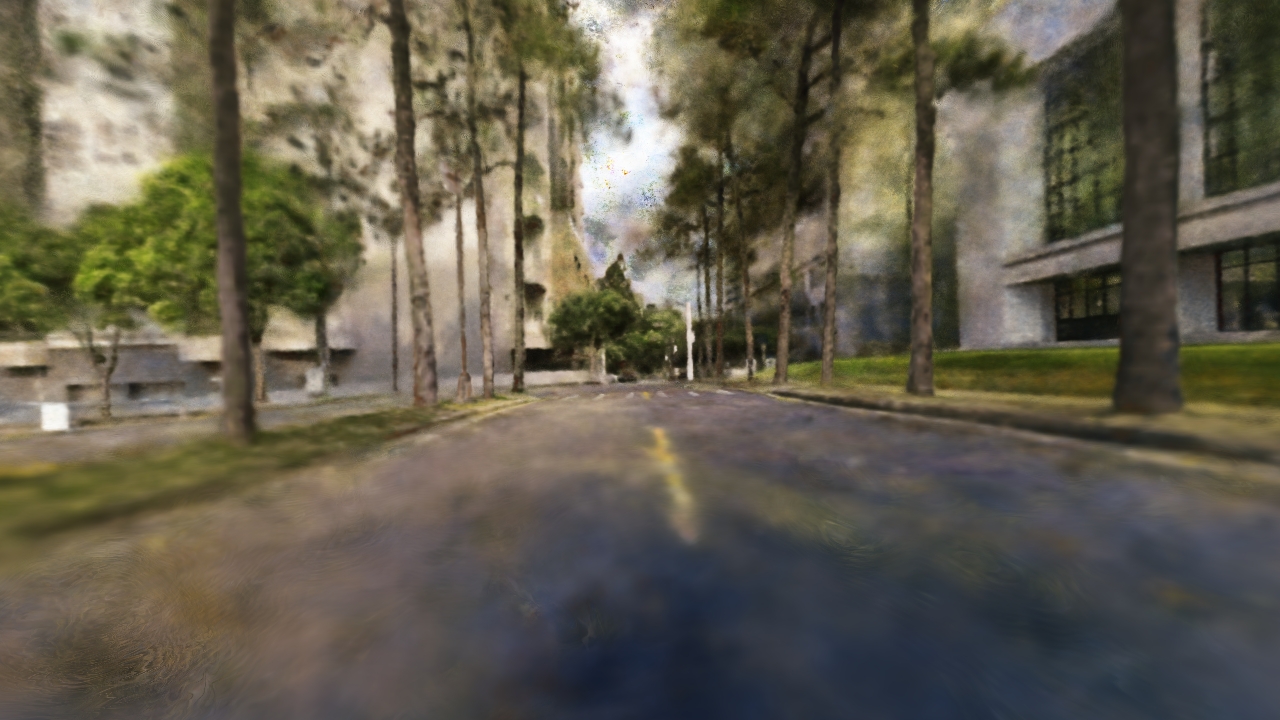}%
    \hfill
    \includegraphics[width=0.16\textwidth, height=0.08\textheight]{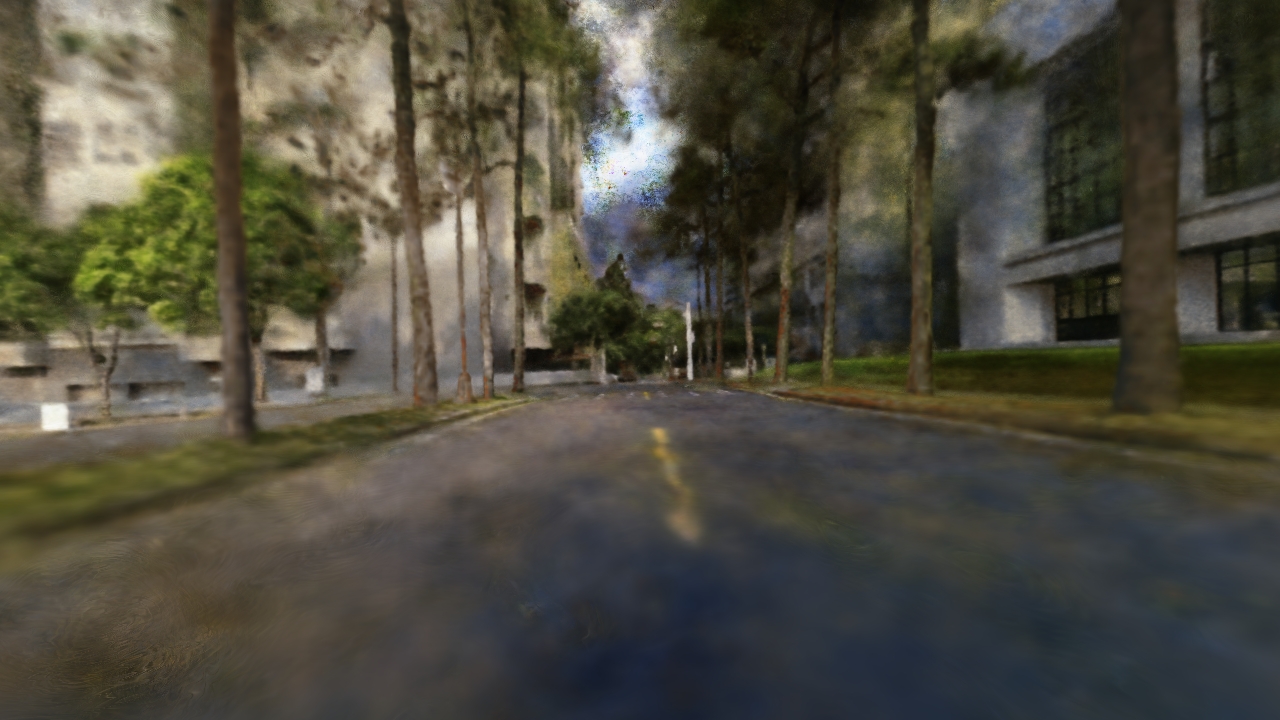}%
    \hfill
    \includegraphics[width=0.16\textwidth, height=0.08\textheight]{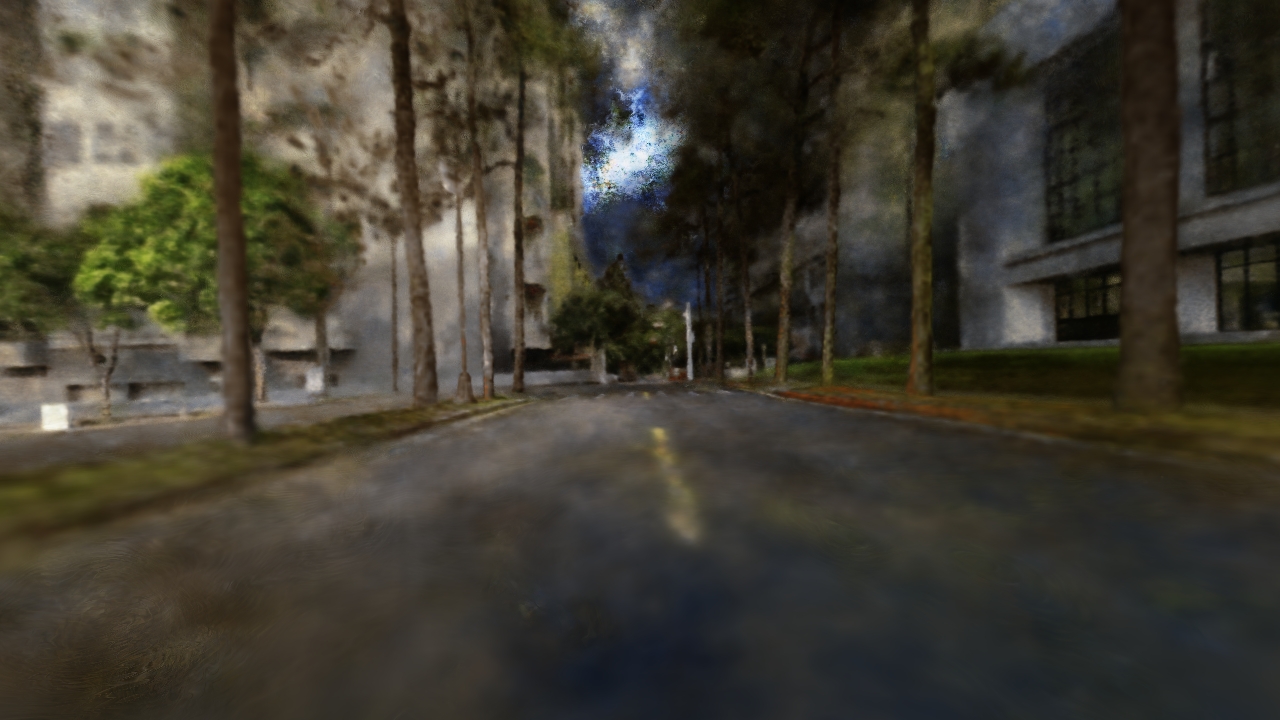}%
    \hfill
    \includegraphics[width=0.16\textwidth, height=0.08\textheight]{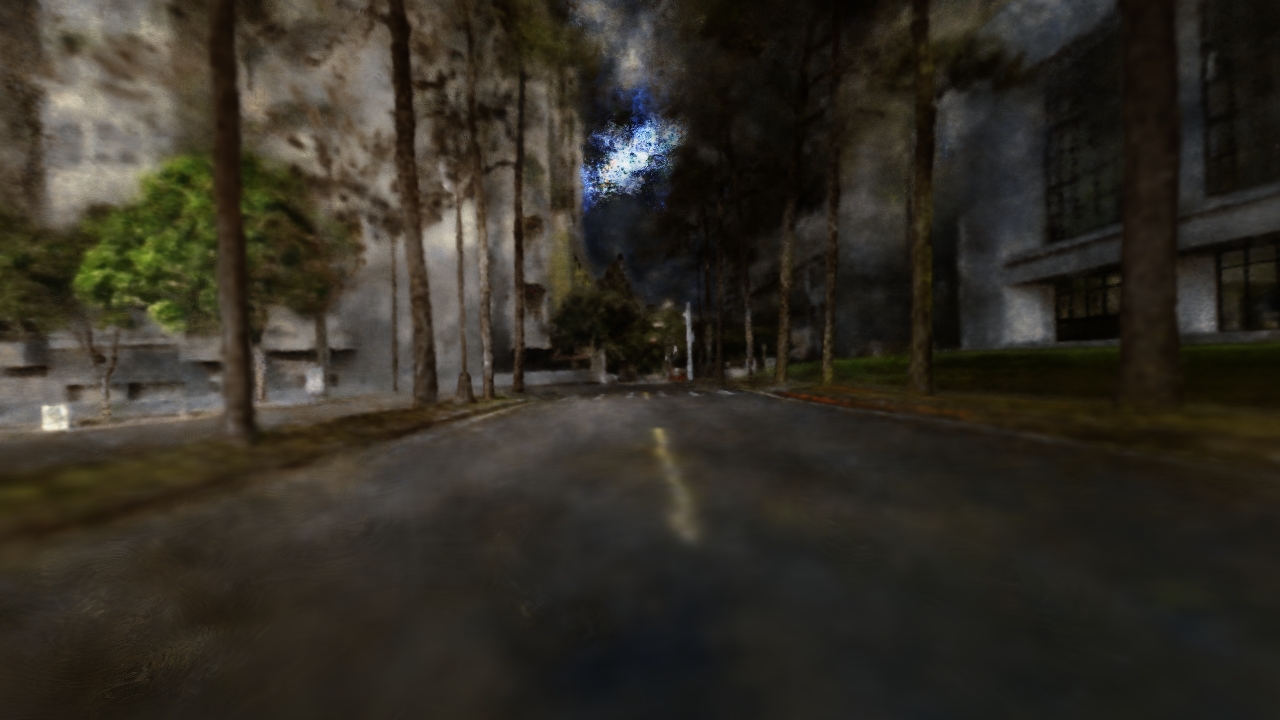}%
    \hfill
    \vspace{0.01\textheight}
    \begin{tikzpicture}
        \draw [->, line width=2pt] (0,0) -- (0.8\textwidth,0);
        \node[anchor=east] at (0,0) {Day};
        \node[anchor=west] at (0.8\textwidth,0) {Night};
    \end{tikzpicture}
    \caption{
    DriveEnv-NeRF enables the interpolation of lighting conditions between two appearance embeddings, which allows generation of a spectrum from a daylight view to a nighttime view. This rendering capability, achieved through the use of appearance embedding, enriches the training and validation process for DRL policies by offering diverse and realistic scenes.}
    
\end{figure*}

\appendix

\input{sections/X.supplementary}
\end{document}

%% file: sections/0.abstract.tex
\begin{abstract}
    In this study, we introduce the DriveEnv-NeRF framework, which leverages Neural Radiance Fields (NeRF) to enable the validation and faithful forecasting of the efficacy of autonomous driving agents in a targeted real-world scene.  Standard simulator-based rendering often fails to accurately reflect real-world performance due to the sim-to-real gap, which represents the disparity between virtual simulations and real-world conditions. To mitigate this gap, we propose a workflow for building a high-fidelity simulation environment of the targeted real-world scene using NeRF. This approach is capable of rendering realistic images from novel viewpoints and constructing 3D meshes for emulating collisions. The validation of these capabilities through the comparison of success rates in both simulated and real environments demonstrates the benefits of using DriveEnv-NeRF as a real-world performance indicator. Furthermore, the DriveEnv-NeRF framework can serve as a training environment for autonomous driving agents under various lighting conditions. This approach enhances the robustness of the agents and reduces performance degradation when deployed to the target real scene, compared to agents fully trained using the standard simulator rendering pipeline.
    \noindent Please note that more results can be found on our project page:\\ \url{https://github.com/muyishen2040/DriveEnvNeRF}.
\end{abstract}

%% file: sections/1.introduction.tex
\section{Introduction}
\label{sec::introduction}
Autonomous driving is a complex task involving real-time decision-making in intricate environments. Although various methods for learning driving policies in virtual environments have been proposed, they often exhibit performance degradation when applied to real-world scenarios. This degradation frequently stems from the discrepancy between simulated and real-world domains in both visual perception and control dynamics, commonly referred to as the sim-to-real gap~\cite{bojarski2016end,pan2017agile,zhang2016query,amini2020learning, kendall2019learning, kiran2021deep, pan2017virtual}. The sim-to-real gap encompasses a wide range of aspects, including perceived observations, environmental conditions, as well as vehicle's physical dynamics. Each of these aspects can affect the practical deployment behavior of the agent and its interaction with real-world objects. As a result, forecasting such discrepancies poses a crucial challenge, particularly when attempting to predict the deployment behavior or performance during the transition from simulated environments to the real world. Therefore, correctly evaluating the real-world performance of a policy during the training phase in a simulated environment is a formidable task. Despite previous endeavors that have attempted to leverage validation environments to anticipate possible outcomes, they often fail to accurately predict actual performance~\cite{kadian2020sim2real,osinski2020simulation}. Moreover, validating the efficacy of trained driving policies directly in real-world scenarios is time-consuming, labor-intensive, and potentially expensive. An appropriate method to address the need for forecasting and validating real-world performance in simulated environments is thus necessary, especially for paradigms concentrating on virtual-to-real transfer learning.

\begin{figure}[t]
\centering
\includegraphics[width=0.4867\textwidth]{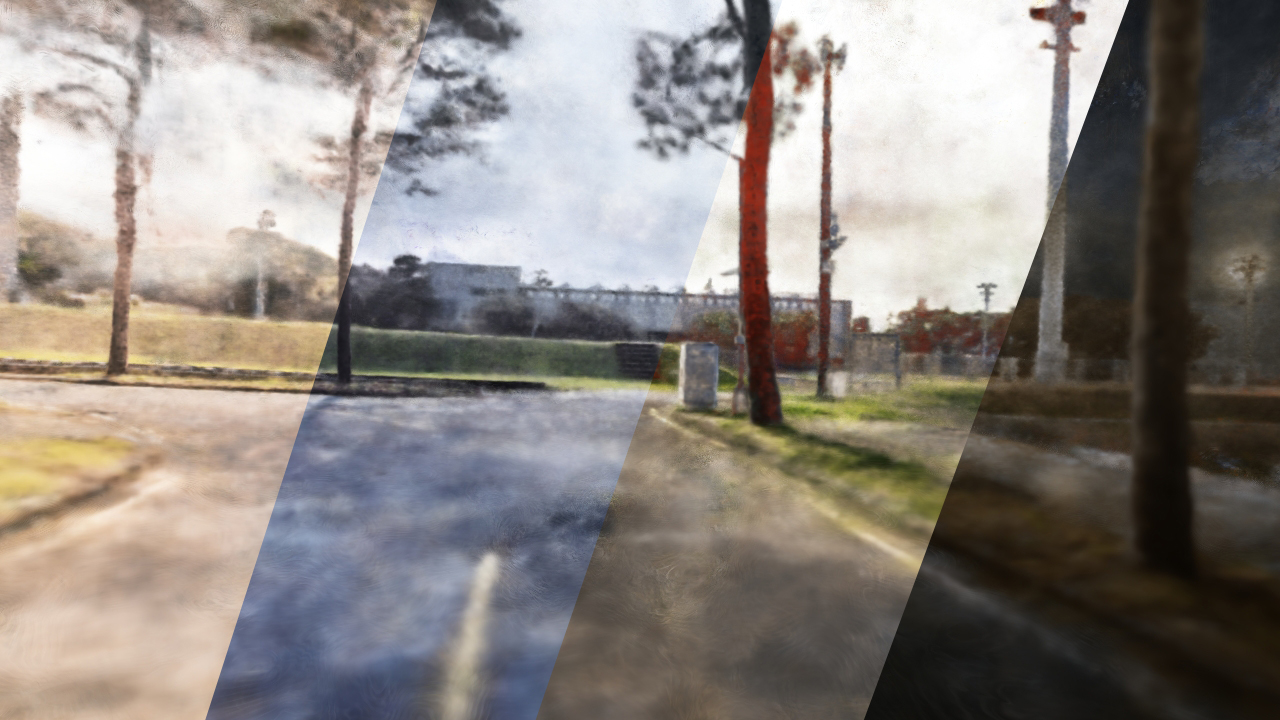}
\caption{Visualization of the scenes rendered by DriveEnv-NeRF for emulating diverse lighting conditions using different setups of the appearance embedding for training policies.}
\label{fig:appemb}
\end{figure}

Photorealistically mirroring real-world environments in a simulator that accurately captures the appearance of the to-be-deployed environment might offer a potential solution to mitigate such a gap. However, reconstructing photorealistic digital twin objects and scenes in virtual environments can be expensive. Factors such as challenging lighting and shadowing conditions, as well as dynamics, introduce complexity to the simulation process. A number of prior research endeavors from different perspectives have been put forth to tackle the virtual-to-real gap, including domain randomization~\cite{tobin2017domain,zhao2020towards, wang2020reinforcement,zhu2023transfer}, domain adaptation~\cite{bousmalis2017unsupervised,hoffman2018cycada,bewley2019learning}, knowledge distillation~\cite{rusu2015policy,traore2019continual,sautier2022image,kothandaraman2021domain,ng2000algorithms}, meta-learning~\cite{wang2016learning,arndt2020meta,liaw2017composing}, learning with disturbances~\cite{wang2020reinforcement}, and others~\cite{zhu2023transfer,lu2019transfer,pinto2017robust}. In addition, some researchers have attempted to address the issue from a different angle by enhancing the simulator's fidelity through traditional or data-driven rendering~\cite{shah2018airsim,dosovitskiy2017carla,furrer2016rotors,amini2022vista,kim2021drivegan,xia2018gibson}.
While these previous efforts have been demonstrated valuable, the challenges of accurately forecasting a model's real-world performance still remain.

In light of the above, this study introduces a framework named DriveEnv-NeRF, capable of forecasting the performance of a driving agent in a targeted real-world environment through building a \simulationenv{}.
Specifically, the \simulationenv{} replicates a target scene from the real-world environment (referred to as the `\textit{\targetrealscene{}}' or the `\textit{\targetrealenv{}}' hereafter) in which our agent intends to operate by utilizing a Neural Radiance Field (NeRF)~\cite{mildenhall2021nerf, tancik2023nerfstudio} model and a physics simulator.
The NeRF model is trained to fit images gathered from the \targetrealenv{} using a cost-effective camera setup. After being trained, it can reconstruct renderings within the \targetrealscene{}, and its extracted 3D meshes are integrated into a simulator for physics simulations.
This offers insights into the forecasted real-world performance when the agent is deployed, which facilitates adjustment of the training procedure to mitigate the impacts of the domain gap between the simulated and the targeted real-world environments.

To more realistically reflect the \targetrealscene{}, we further utilize the concept of appearance embedding proposed in NeRF-W~\cite{martin2021nerf} as a means of data augmentation. This enables diversifying the lighting conditions in the simulated scene, which allows our agents to be deployed in various conditions such as daytime, dawn, or even in the evening, as illustrated in Fig.~\ref{fig:appemb}. The contributions can be summarized as follows:

\begin{itemize}
\item We introduce the DriveEnv-NeRF framework for building a \simulationenv{}, which enables forecasting the performance of an agent's policy in real scenarios.
\item We employ NeRF to model and reconstruct the \targetrealenv{} and extract 3D meshes from it to provide a realistic emulation of the \targetrealenv{}.
\item We validate the effectiveness of the simulation environment constructed by DriveEnv-NeRF as a performance indicator for the policy trained by deep reinforcement learning (DRL) by conducting real-world experiments.
\item We propose the use of appearance embedding to augment the training effectiveness by altering the lighting conditions to emulate more diverse training conditions.
\end{itemize}

%% file: sections/2.related_works.tex
\section{Related Works}
NeRF has gained considerable attention for its capability of synthesizing novel photo-realistic views with implicit scene representation. The success of NeRF has led to an explosion of subsequent methods addressing quality, speed, accurate scene geometry, and other aspects~\cite{tancik2022block,turki2022mega,muller2022instant,yu2021plenoctrees,sun2022direct,kerbl20233d,wang2021nerf,lin2021barf,reiser2021kilonerf, barron2021mip, barron2022mip, barron2023zip}. While NeRF performs well on static scenes with controlled lighting conditions, it encounters challenges when dealing with image collections from unconstrained real-world scenarios, such as variable weather, lighting, or transient occluders. NeRF-W~\cite{martin2021nerf} attempts to address these issues with appearance embeddings and transient networks. The existence of appearance embedding further enables implicit control of illumination. In addition to photo-realistic rendering, NeRF also allows for the extraction of accurate geometry. Methods such as Marching Cubes~\cite{Lorensen1987MarchingCA} or Poisson Surface Reconstruction~\cite{kazhdan2006poisson} can extract 3D meshes from NeRF, which can be further utilized for physical simulations. The combination of 3D meshes and rendered images presents the possibility of constructing a real-to-sim simulation environment with NeRF~\cite{byravan2023nerf2real, zhou2023nerf, adamkiewicz2022vision}. NeRF2Real~\cite{byravan2023nerf2real} employs NeRF to reconstruct real-world environments in a simulator, which enables novel view synthesis and collision detection. Based on such a simulator, a bipedal robot control policy trained with DRL can be deployed to the real world in a zero-shot manner.

%% file: sections/3.methodology.tex
\section{Methodology: DriveEnv-NeRF Framework}
\begin{figure*}[tp]
    \centering
    \begin{minipage}[b]{0.7\linewidth}
        \centering
        \includegraphics[width=\linewidth]{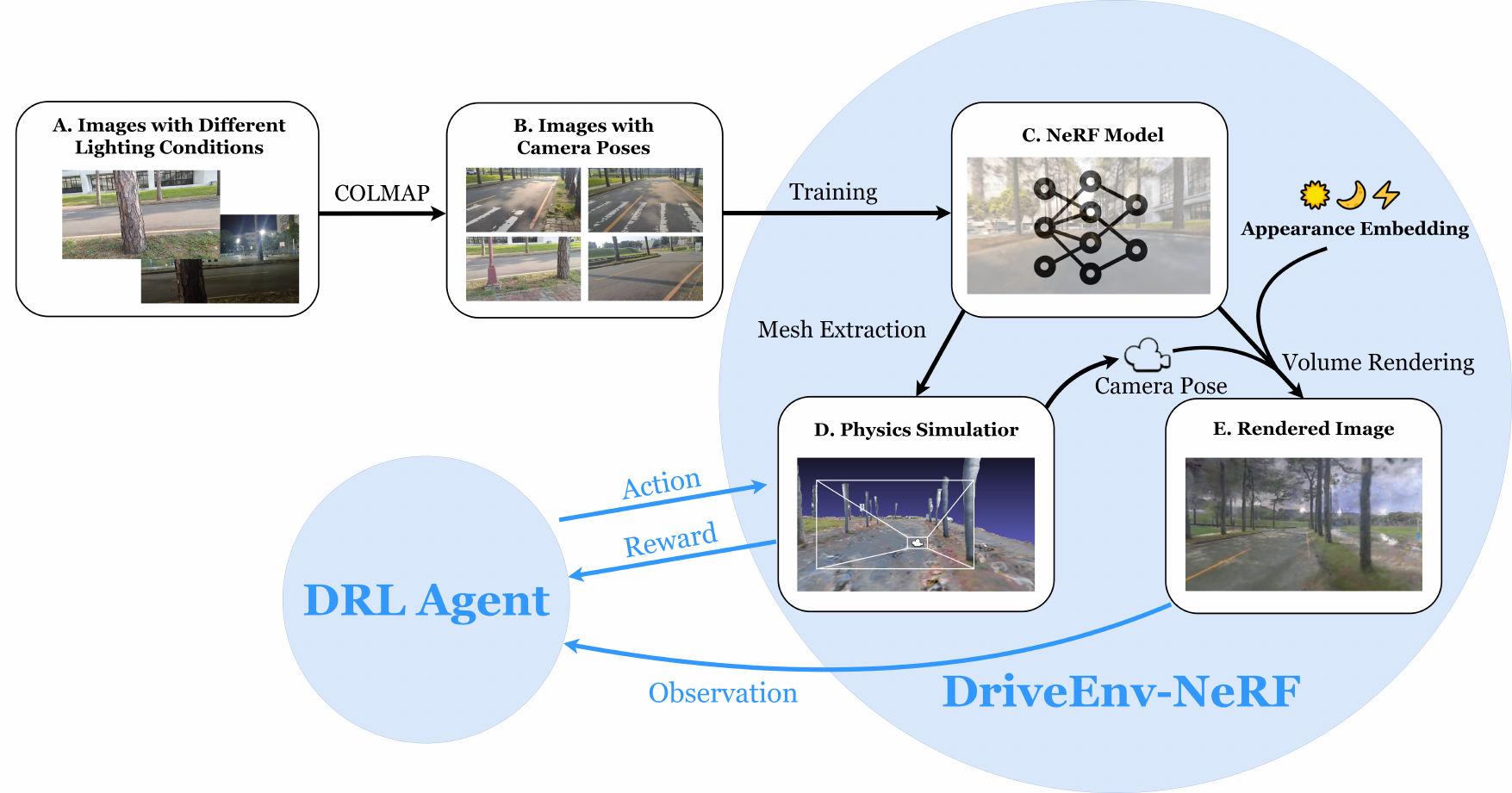}
        \caption{An overview of the DriveEnv-NeRF framework.}
        \label{fig:overview}
    \end{minipage}
    \hfill
    \begin{minipage}[b]{0.25\linewidth}
        \centering
        \begin{subfigure}[b]{\linewidth}
            \centering
            \includegraphics[width=\linewidth]{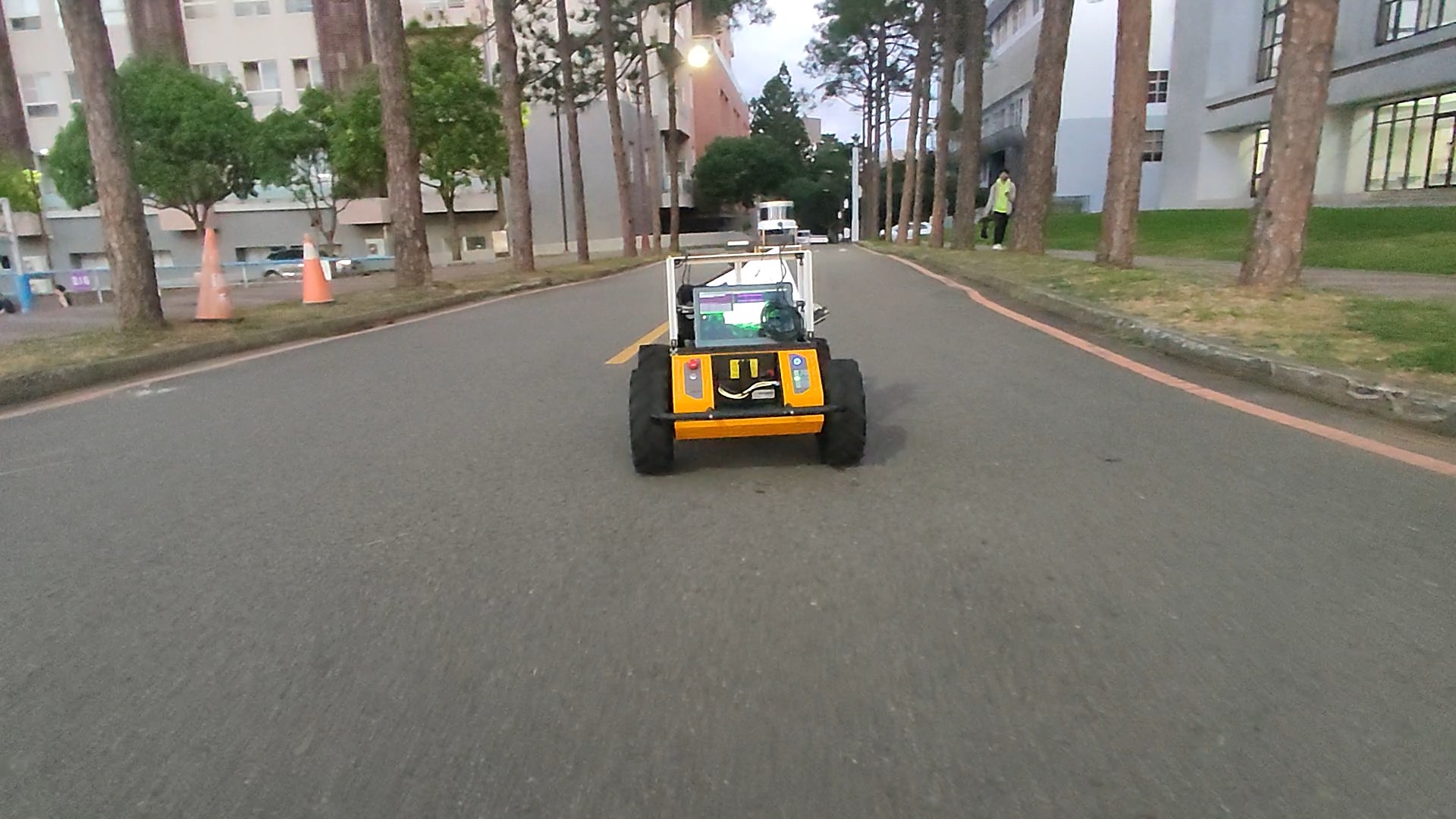}
            \label{fig:top-right}
        \end{subfigure}
        \vspace{-10pt}
        \begin{subfigure}[b]{\linewidth}
            \centering
            \includegraphics[width=\linewidth]{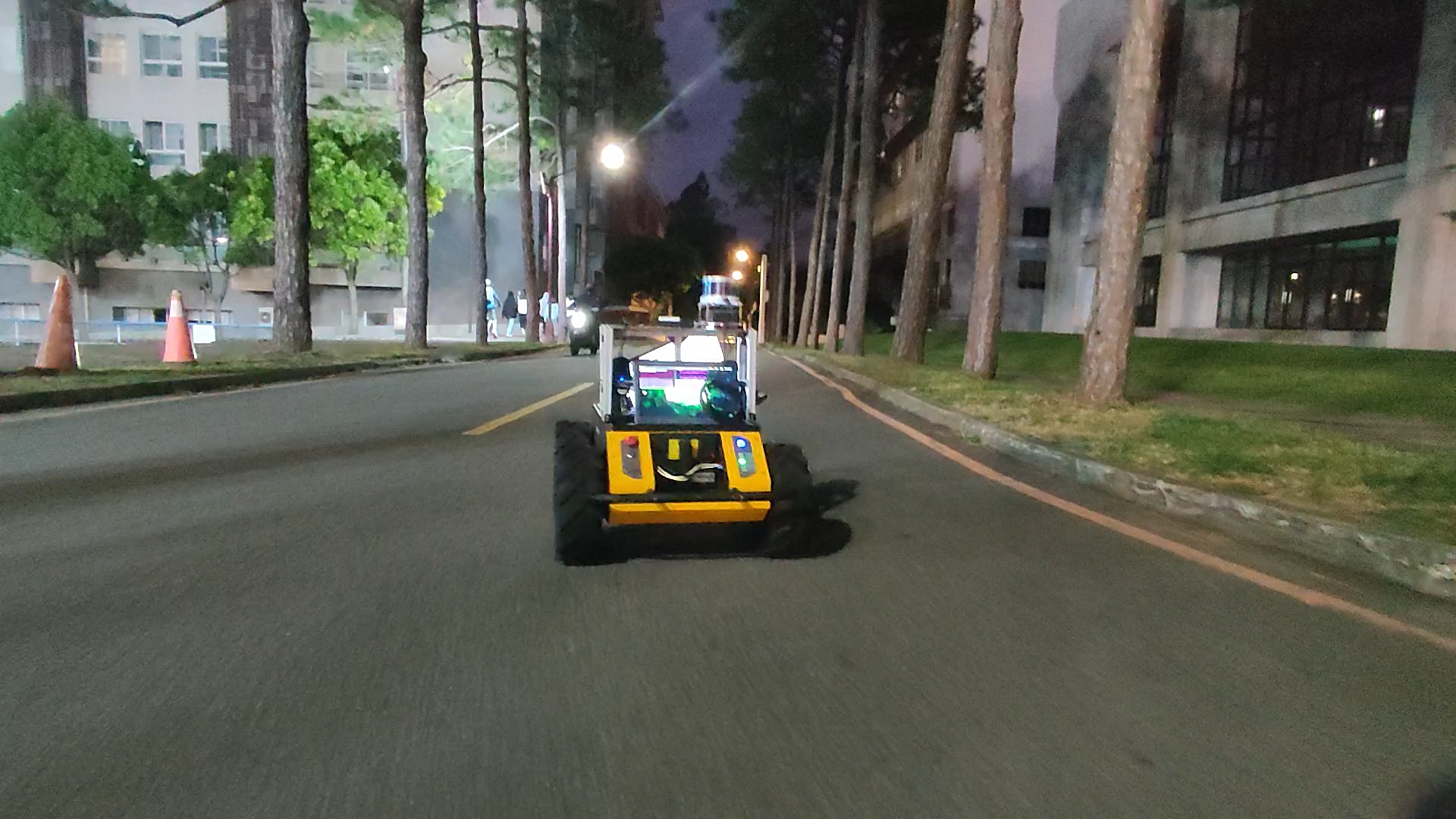}
            \label{fig:bottom-right}
        \end{subfigure}
        \caption{The application of DriveEnv-NeRF shows agent's adaptability to diverse lighting.}
        \label{fig:right-images}
    \end{minipage}
\end{figure*}

\begin{figure*}[tp]
  \centering
  \begin{subfigure}[b]{0.28\textwidth}
    \includegraphics[width=\textwidth]{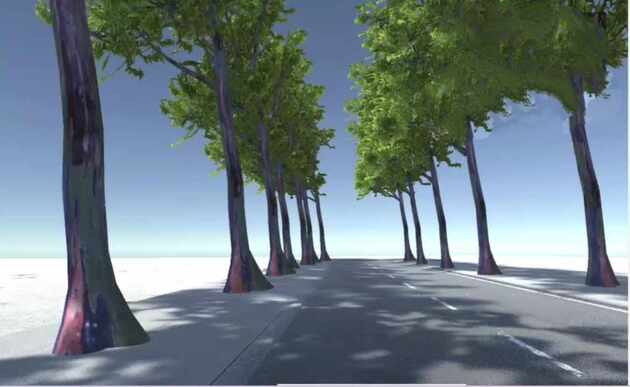}
    \caption{}
  \end{subfigure}
  \hspace{12pt}
  \begin{subfigure}[b]{0.28\textwidth}
    \includegraphics[width=\textwidth]{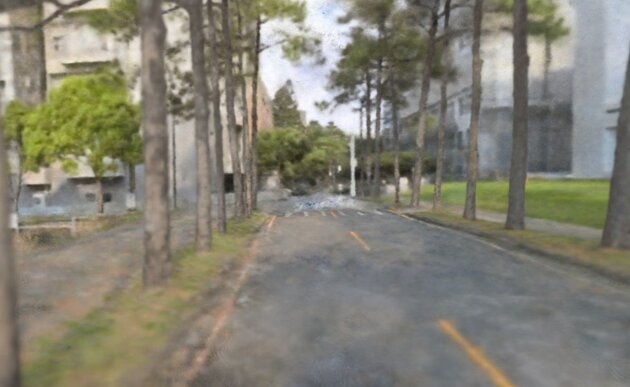}
    \caption{}
  \end{subfigure}
  \hspace{12pt}
  \begin{subfigure}[b]{0.28\textwidth}
    \includegraphics[width=\textwidth]{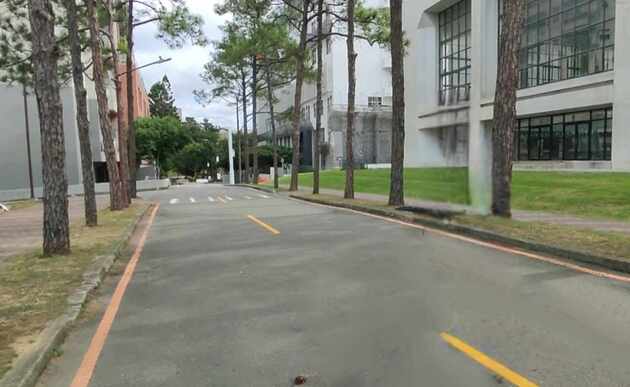}
    \caption{}
  \end{subfigure}
  \caption{Examples of (a) an image rendered by the simulator, (b) a NeRF rendered image, and (c) a real image.}\vspace{-1em}
  \label{fig:diffview}
\end{figure*}

Fig.~\ref{fig:overview} illustrates an overview of the proposed DriveEnv-NeRF framework, which aims to construct a \simulationenv{} that better reflects a real-world deployment scene. To achieve this, DriveEnv-NeRF utilizes images of the \targetrealscene{} to train a NeRF model that serves two purposes. First, the trained NeRF model is utilized to generate novel views for the agent during validation and DRL training. Second, 3D meshes of the scene are extracted from the trained NeRF model to emulate physical presence. As a result, DriveEnv-NeRF is able to reflect the environment appearance for visual observations and utilize the derived 3D meshes in the simulated environments to model collisions. The components of DriveEnv-NeRF are detailed as follows.

\subsection{Construction of the Simulation Environment}

The DriveEnv-NeRF framework encompasses four main stages: (1) Data Collection and Preprocessing, (2) Scene Reconstruction, (3) Physics Simulation, and (4) Hardware Alignment. Each of these stages is elaborated on as follows.

\textbf{Data Collection and Preprocessing.}
To train the NeRF model, videos of the \targetrealscene{} are captured under various lighting conditions, such as different times of the day and weather scenarios. Image frames are then extracted at regular intervals to ensure high-quality and unobstructed views. These frames are processed using the structure-from-motion (SfM) process with COLMAP~\cite{schonberger2016structure,schonberger2016pixelwise} to establish correspondences, which assist in determining the camera frames' extrinsic and intrinsic parameters for NeRF training.

\textbf{Scene Reconstruction.}
To reconstruct the \simulationenv{} for DriveEnv-NeRF, the NeRF model is first trained to generate novel view renderings, which are then utilized to derive the 3D meshes of the \targetrealscene{}. The rendered images serve as input observations for the agent, while the 3D meshes are employed for collision detection. The NeRF model further incorporates an appearance embedding, which enables the representation of a scene under various lighting conditions. During inference, the appearance embedding can be specified to render images with different illuminations.

\textbf{Physics Simulation.}
The \simulationenv{} is built upon the Unity engine~\cite{unity}, a game engine with physics simulation capabilities for collision detection and 3D object management. The reference frames of the NeRF model and the 3D meshes in the simulator are aligned, which enables translation of coordinates between the simulated frames and the NeRF-rendered frames. Another key feature of this environment is its support for the incorporation of various 3D objects, such as pedestrians and obstacles, into the rendered scenes to enrich the environmental details and complexity.

\textbf{Hardware Alignment.}
To faithfully capture real-world performance using the \simulationenv{} built by DriveEnv-NeRF, it is essential to align the agent's configuration with real-world conditions. This involves adjusting parameters such as camera intrinsics, the camera's height and angle relative to the ground and the vehicle, as well as aligning the vehicle's speed and turning angle. Other pertinent factors are taken into account for distinct use cases.

\subsection{Validation and Training of DRL Agents}
Once the \simulationenv{} is built, the training and evaluation of driving policies can be fully conducted within it. At each timestep, the camera coordinates of the driving agent in the simulator are transformed to the corresponding coordinates in the NeRF model for rendering an image with NeRF. This image is then used as the agent's observation. Subsequently, the action selected by the agent is sent to the simulator, which executes it to simulate the agent's movements. This \simulationenv{} can be used to both train and validate the agent's driving policy with different settings. More specifically, the road layout and lighting conditions can be fixed to serve as a standardized and reproducible validation condition. If more complex environments are desired for validation, DriveEnv-NeRF allows alternation of the appearance embedding to evaluate the policy's generalizability over different lighting or weather conditions, as depicted in Fig.~\ref{fig:appemb}. This feature is crucial in training a driving policy as it reduces the negative impacts of domain discrepancy caused by factors such as weather and lighting. Moreover, static or dynamic obstacles can be generated using the simulator.
As a result, the \simulationenv{} built with the DriveEnv-NeRF framework can serve as a benchmark for validating driving policies trained using different procedures through the emulation of these policies within the environment.
By leveraging the realistic images rendered by NeRF, as illustrated in Fig.~\ref{fig:diffview}, the trained DRL policies can be validated within the reconstructed \simulationenv{} using success rate or other types of metrics to forecast their performance in the \targetrealscene{}.

%% file: sections/4.experimental_results.tex
\section{Experimental Results}

\subsection{Experimental Setup}

\textbf{Tasks.}
Two different navigation tasks, namely \textit{Straight Road} and \textit{Single Right Turn}, are utilized to demonstrate the effectiveness of our DriveEnv-NeRF framework for forecasting the performance of autonomous driving policies in the real world. In these tasks, the DRL agent is expected to safely navigate the road without deviating from the boundaries.

\textbf{Validation Environments.}
Two validation environments, (1) the baseline simulator and (2) the \simulationenv{} built by DriveEnv-NeRF, are utilized to assess their effectiveness in predicting the model's performance in the real world. In the baseline simulator, the agent can only observe the manually-built mesh of the scene. DriveEnv-NeRF, on the other hand, allows the agent to observe the renderings produced by NeRF. The DRL agent can be evaluated under an appearance embedding not encountered during training.

\textbf{DRL Training Setup.}
To validate the effectiveness of our proposed environment, we employ the Soft Actor-Critic (SAC) algorithm~\cite{haarnoja2018soft} as the DRL policy. In addition, the agent is trained from scratch with raw RGB observations in an end-to-end manner, without relying on prior datasets for pretraining. This specific setup aims to highlight the benefits offered by DriveEnv-NeRF. All training settings and model parameters are fixed across experiments. In our comparisons, we consider three baseline approaches described as follows:

\begin{itemize}
    \item \textit {Simulator:} This baseline takes input image observations generated by the standard simulator rendering pipeline.

    \item \textit {DriveEnv-NeRF w/o appearance embedding:} The agent receives NeRF renderings as inputs. Nevertheless, the appearance embedding is kept constant during training.
    
    \item \textit {DriveEnv-NeRF w/ appearance embedding:} This is the default DriveEnv-NeRF with varying appearance embeddings between episodes for diverse appearances.
\end{itemize}

\textbf{Evaluation Metric.}
In the experiments, the success rate is employed to evaluate the performance of the trained DRL agents. An episode is considered successful if the DRL agent reaches its destination without deviating from the road boundaries, causing collisions, or exceeding the time limit.

\subsection{Quantitative Results}
\vspace{-0.5em}

\begin{table}[htbp]
\centering
\captionsetup{skip=10pt} 
\caption{
The success rate of the agents trained under different baseline setups. `A.E.' stands for appearance embedding. Each experiment is performed with ten independent runs.
} 
\label{tab:results}
\resizebox{\columnwidth}{!}{%
\begin{tabular}{@{} l c c c @{}}
\toprule
\multicolumn{4}{c}{\textbf{Task 1 - Straight Road}} \\
\toprule

Baselines & Training & Validation & Real-World Testing \\
\midrule
Simulator           & 0.94 & 0.57 & 0.00 \\
DriveEnv-NeRF w/o A.E.   & 1.00 & 0.19 & 0.00 \\
DriveEnv-NeRF w/ A.E.    & \textbf{0.99} & \textbf{0.98} & \textbf{0.80} \\
\bottomrule
\toprule
\multicolumn{4}{c}{\textbf{Task 2 - Single Right Turn}} \\
\toprule
Baselines & Training & Validation & Real-World Testing \\
\midrule
Simulator           & 0.73 & 0.54 & 0.00 \\
DriveEnv-NeRF w/o A.E.   & 1.00 & 0.95 & 0.67 \\
DriveEnv-NeRF w/ A.E.    & \textbf{1.00} & \textbf{1.00} & \textbf{0.83} \\
\bottomrule
\end{tabular}
}
\end{table}

Table~\ref{tab:results} demonstrates the performances of the agents in different environments. The \textit{Simulator} setup trains and validates the agent entirely using the standard simulator rendered image. In contrast, the \textit{DriveEnv-NeRF} setup trains and validates the agent in the simulation environment built by DriveEnv-NeRF. One of the \textit{DriveEnv-NeRF} setup utilizes appearance embeddings, while the other does not. In the \textit{Real-World Testing} environment, we conduct experiments on a ClearPath Husky Unmanned Ground Vehicle (UGV) equipped with a ZED-2 camera, as depicted in Fig~\ref{fig:right-images}. 
Please note that in this study, the agent use raw image only, and does not rely on the depth estimation capability of ZED-2.

\textbf{Results of Straight Road Task.}
The results indicate that despite achieving a high success rate during assessments within the training simulated environment for all three cases, only the policies trained under the NeRF environment with varying appearance embedding demonstrate the capability to reach the end-point without deviating from the straight road. These policies exhibit superior adaptability to different lighting conditions in the real-world scenarios and can achieve approximately an 80\% success rate when driving through a straight road. In contrast, the other baselines exhibit a significantly higher probability of deviating from the road.

\textbf{Results of Single Right Turn Task.}
The task of executing a right turn poses greater challenges when applied to the target real world. It can be observed that only the DRL models trained under NeRF environments can successfully complete the task. Similar to the previous task, providing the scene's various light conditions with appearance embedding during training can result in a more robust driving policy.

\subsection{Sim-to-Real Gap Analysis}

In both straight driving and right turn tasks, success rates below 0.6 in the \textit{Validation} environments consistently lead to failure in \textit{Real-World Testing} environments. A model that outperforms others in the \simulationenv{} built by DriveEnv-NeRF still maintains its advantage in the real world. However, the results indicate that high success rates in the simulator do not guarantee similar performance in real-world scenarios due to the sim-to-real domain gap. These findings validate that DriveEnv-NeRF can serve as an effective predictor for real-world performance and can be used to forecast the relative performance of different policies.

Although DriveEnv-NeRF demonstrates performance alignment with the real world, a subtle disparity persists due to the inherent challenges in bridging the sim-to-real gap. Achieving precise physical calibration between simulation and real-world scenarios presents a significant obstacle. Fluctuations in camera extrinsics caused by UGV vibration and inaccuracies in UGV wheel encoders during slipping may further impact the performance of the DRL policy. Such discrepancies can complicate the application of experiences gained in the simulation environment to real-world scenarios.

%% file: sections/5.discussion_and_conclusion.tex
\section{Discussion and Conclusion}
In this work, we presented DriveEnv-NeRF, a simulation framework using NeRFs to validate and forecast the real-world performance of autonomous driving agents. Our method adjusts appearance embeddings to reflect varying lighting conditions, and the results demonstrate its effectiveness as the success rates in simulated and real-world settings align. DriveEnv-NeRF also serves as a beneficial training platform for enhancing the adaptability and robustness of policies through the alteration of environmental appearances.

While DriveEnv-NeRF demonstrates promising results, opportunities for further refinement exist in novel view synthesis methods and simulator selection. Recent advancements introduced 3D Gaussian Splatting (3DGS)~\cite{kerbl20233d} for novel view synthesis, which offered realistic rendering at significantly faster speeds. 3DGS facilitates direct collision detection on point clouds and can potentially be enhanced with tailored mesh extraction methods such as SuGaR~\cite{guedon2023sugar}. DriveEnv-NeRF, built on top of the Unity engine, faces challenges such as integration with practical robotics due to the gap between hardware simulation and real-world implementation. To address these, a potential avenue is shifting to Isaac Sim~\cite{nvidia_isaac_sim} together with ROS 2~\cite{macenski2022robot} and Isaac ROS~\cite{nvidia_isaac_ros}, which offer better tool integration and simulation fidelity.

%% file: sections/6.acknowledgement.tex
\section*{Acknowledgement}

The authors gratefully acknowledge the support from the National Science and Technology Council (NSTC) in Taiwan under grant numbers MOST 111-2223-E-007-004- MY3, as well as the financial support from MediaTek Inc., Taiwan. The authors would also like to express their appreciation for the donation of the GPUs from NVIDIA Corporation and NVIDIA AI Technology Center (NVAITC) used in this work. Furthermore, the authors extend their gratitude to the National Center for High-Performance Computing (NCHC) for providing the necessary computational and storage resources.

%% file: sections/X.supplementary.tex
\label{supplementary}

\subsection{An Extended Review of the Related Work}
\textbf{NeRF.}
Nerfstudio~\cite{tancik2023nerfstudio} is a modular framework for NeRF development that allows rapid exploration and implementation of new concepts. Within this framework, Nerfacto is the default model, which combines a number of state-of-the-art (SOTA) methods, namely Mip-NeRF 360~\cite{barron2022mip}, Instant-NGP~\cite{muller2022instant}, NeRF-W~\cite{martin2021nerf}, NeRF$--$~\cite{wang2021nerf}, Ref-NeRF\cite{verbin2022ref}, etc.NeRF~\cite{verbin2022ref}, etc.

\textbf{Training DRL with NeRF for Robotics.}
SPARTN~\cite{zhou2023nerf} augments expert data in imitation learning scenarios by leveraging NeRFs to generate new viewpoints based on perturbed expert camera poses. Subsequently, the associated expert actions are adjusted based on the known camera-to-end-effector transform. The study in~ \citet{adamkiewicz2022vision} establishes a robot navigation pipeline based on a pre-trained NeRF for representing a 3D environment. Its trajectory optimization algorithm is designed to circumvent high-density regions of the pre-trained NeRF model, which ensures the generation of collision-free trajectories. In addition, a pose filtering mechanism leveraging NeRF's renderings is incorporated to estimate the robot's pose and velocities, which further improves the overall accuracy of the estimated robot state.

Apart from directly using the generated renderings, another line of research~\cite{li20223d,driess2022reinforcement,shim2023snerl} explores latent-conditioned NeRF to train an encoder-decoder model that learns the 3D representations of a scene. The intermediate 3D representations generated by the encoder are utilized in downstream DRL tasks, which enhances the model's learning capabilities by incorporating pre-trained information. This 3D-aware neural implicit representation derived from NeRF empowers the DRL agent to generalize across diverse scene geometries. Nevertheless, these methods are orthogonal to this study.

\textbf{NeRF-based Simulators.}
Concurrent to our study, several previous works~\cite{wu2023mars, yan2024oasim, Yang_2023_CVPR, tonderski2023neurad} have attempted to build autonomous self-driving simulators using NeRF as a foundation. These approaches focus more on the development of realistic simulators for general datasets, including the insertion of dynamic objects and the generation of corner cases. However, our methodology diverges from these studies as our primary focus is on replicating specific target scenes from the real-world environment where the robot is intended to be deployed, rather than creating a general-purpose simulator.

\subsection{Details of the Experimental Setups}
\textbf{Target Real Scenes.}
The experiments focus on two specific real-world environments: a straight road and a right turn, measuring 30 meters and 60 meters in length, respectively. Approximately 500 photographs were captured to train the NeRFs employed in our proposed pipeline. These images were taken under various weather conditions to ensure a diverse representation of environmental appearances in the NeRF models.

\textbf{DRL Environment Setup.}
The DRL agent receives a survival reward of 0.1 at each timestep. On the other hand, a penalty of -4.0 is imposed for collision with obstacles, and -5.0 points are deducted for moving out of boundaries or exceeding the time limit. The observation of the agent consists of four stacked RGB frames, where each frame is individually rendered by NeRF with a resolution of 90 x 160.

\textbf{NeRF Training Setup.}
In our experiments, we utilize Nerfacto~\cite{tancik2023nerfstudio} for the reconstruction of our targeted real-world scenes. The training process involves 200,000 iterations with a batch size of 4,096, and the training can be completed in approximately 10 minutes on an NVIDIA RTX Titan GPU.

\textbf{Appearance Embedding.}
To generate appearance embeddings for evaluating and training our driving policy within the proposed framework, we sample a diverse range of lighting conditions by applying randomness to each dimension of the embedding. These embeddings are subsequently selected manually based on their ability to realistically represent various environmental lighting scenarios. The dimension of the appearance embedding used in the experiment is 32.

\textbf{Real-World Experimental Setup.}
We deploy a trained driving policy on the Clearpath Husky UGV, which receives commands from a Jetson Xavier NX. The DRL model is executed on a laptop equipped with a GeForce RTX 4080 GPU, and the derived actions are forwarded to the Jetson Xavier NX, which transforms the action commands into control signals for the Husky UGV for real-time execution.

\subsection{Extended Discussion and Conclusion}

In addition to the potential future improvements on novel view synthesis methods using 3DGS~\cite{kerbl20233d}, there is also room for enhancement in the deployment process of the driving agent by leveraging dedicated simulators such as Isaac Sim~\cite{nvidia_isaac_sim}. A software transition from ROS 1 to ROS 2~\cite{macenski2022robot}, further augmented by the integration of Isaac ROS~\cite{nvidia_isaac_ros}, can facilitate improved real-time performance. Moreover, ROS controller nodes can be encapsulated within containers to simplify deployment across diverse hardware platforms. To facilitate further exploration of our project, we've shared a portion of our implementation code in Isaac Sim and Isaac ROS on the project page. This page will be regularly updated as we develop more open-source implementations. In conclusion, DriveEnv-NeRF paves the way for cost-effective validation by facilitating the replication of real-world scenes and enhancing continuous testing of driving policies. Employing DriveEnv-NeRF in common real-world failure scenarios can strengthen the testing pipeline by faithfully mirroring these challenging real-world situations. Future research directed towards refining and expanding the capabilities of DriveEnv-NeRF to achieve a more precise alignment with real-world performance presents a promising avenue for further investigation and advancement in the field.